\documentclass{article} 
\usepackage{iclr2025_conference,times}

\usepackage{amsmath,amsfonts,bm}

\def\eqref#1{equation~\ref{#1}}

\def\1{\bm{1}}

\DeclareMathAlphabet{\mathsfit}{\encodingdefault}{\sfdefault}{m}{sl}
\SetMathAlphabet{\mathsfit}{bold}{\encodingdefault}{\sfdefault}{bx}{n}

\usepackage{hyperref}
\usepackage{url}
\usepackage{makecell}
\usepackage{graphicx}
\usepackage{multirow}

\newcommand{\dataname}{\textsc{FaultyMath}}
\newcommand{\solver}{\textsc{Blind Solver}}
\newcommand{\thinker}{\textsc{Logical Thinker}}

\title{From Blind Solvers to Logical Thinkers: Benchmarking LLMs' Logical Integrity on Faulty Mathematical Problems}

\author{
    A M Muntasir Rahman$^{\sharp}$\thanks{Equal contribution.}, 
    Junyi Ye$^{\sharp}$\footnotemark[1], 
    Wei Yao$^{\sharp}$,
    Sierra S. Liu$^{\dagger}$, 
    Jesse Yu$^{\dagger}$, 
    Jonathan Yu$^{\dagger}$, \\
    \textbf{Wenpeng Yin$^{\P}$, and
    Guiling Wang$^{\sharp}$} \\
    $^{\sharp}$New Jersey Institute of Technology
    $^{\dagger}$Millburn High School
    $^{\P}$Pennsylvania State University
    \\
    \texttt{\{ar238,jy394,wy95,guiling.wang\}@njit.edu} \\
    \texttt{\{Sierra.Sijia.Liu,jesse.gift.yu,jon.sunny.yu\}@gmail.com} \\
    \texttt{wenpeng@psu.edu}
}

\iclrfinalcopy 
\begin{document}

\maketitle

\begin{figure}[!ht]
\centering
\includegraphics[width=1.0\columnwidth]{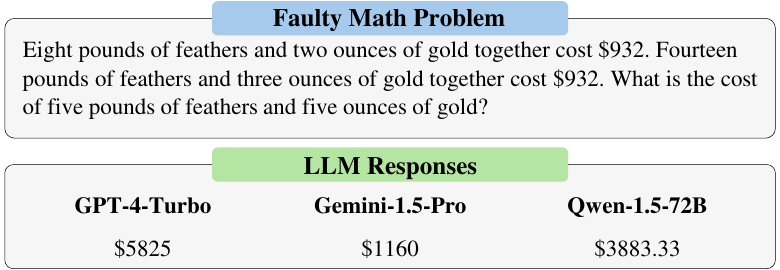} 
\caption{A faulty math problem and answers by popular LLMs. }
\label{fig:false_math_problem_example}
\end{figure}
\begin{abstract}
Consider the math problem: ``Lily received 3 cookies from her best friend yesterday and ate 5 for breakfast. Today, her friend gave her 3 more cookies. How many cookies does Lily have now?'' Many large language models (LLMs) in previous research approach this problem by calculating the answer ``1'' using the equation ``3 - 5 + 3.'' However, from a human perspective, we recognize the inherent flaw in this problem: Lily cannot eat 5 cookies if she initially only had 3. This discrepancy prompts a key question: Are current LLMs merely \solver~that apply mathematical operations without deeper reasoning, or can they function as \thinker~capable of identifying logical inconsistencies?

To explore this question, we propose a benchmark dataset, \dataname, which includes faulty math problems of rich diversity: i) multiple mathematical categories, e.g., algebra, geometry, number theory, etc., ii) varying levels of difficulty, and iii) different origins of faultiness—ranging from violations of common sense and ambiguous statements to mathematical contradictions and more. We evaluate a broad spectrum of LLMs, including open-source, closed-source, and math-specialized models, using \dataname~across three dimensions: (i) How accurately can the models detect faulty math problems without being explicitly prompted to do so? (ii) When provided with hints—either correct or misleading—about the validity of the problems, to what extent do LLMs adapt to become reliable \thinker? (iii) How trustworthy are the explanations generated by LLMs when they recognize a math problem as flawed? Through extensive experimentation and detailed analysis, our results demonstrate that existing LLMs largely function as \solver~and fall short of the reasoning capabilities required to perform as \thinker\footnote{Data\&Code: \url{https://github.com/JunyiYe/FaultyMathProblem}}.

\end{abstract}

\section{Introduction}

Our exploration begins with the example presented in Figure \ref{fig:false_math_problem_example}, where a logically flawed math problem still leads prominent large language models (LLMs) to produce various numerical answers. This outcome is surprising, as we would expect an intelligent AI system not only to provide correct solutions to valid math problems but also to identify when a problem is logically inconsistent and therefore should not yield any answer, even without explicit hints about its flawed nature. This observation prompts us to re-examine the current development trajectory of LLMs for mathematical problem-solving \citep{DBLPAhnVLLZY24,hendrycks2021measuring,DBLPImaniD023}, which generally assume that the provided math problems are valid and require only computational solutions, without considering the possibility of logical flaws. This raises a crucial question: Are contemporary LLMs merely \solver~that execute mathematical operations without deeper reasoning, or can they evolve into \thinker~capable of detecting logical inconsistencies?

To address this question, we introduce a new dataset, \dataname, which incorporates three dimensions of diversity: (i) a range of mathematical categories, including algebra, geometry,  number theory, etc., (ii) varying levels of difficulty, and (iii) a rich variety of falsehoods, such as violations of common sense, insufficient information,  mathematical contradictions, etc.

The creation of \dataname~was a multi-step process. Initially, we recruited participants from the American Mathematics Competitions (AMC) to manually generate faulty math problems. However, this led to a dataset that was disproportionately skewed toward specific mathematical categories and types of logical flaws. To create a more balanced and comprehensive dataset, we developed a three-stage pipeline: (i) We used GPT-4 \citep{achiam2023gpt} to modify valid problems from the MATH dataset \citep{hendrycks2021measuring} into flawed ones, generating 2,000 potential faulty math problems; (ii) we then prompted GPT-4 to perform self-evaluation, identifying 635 of these as likely faulty; (iii) finally, human annotators reviewed these problems, confirming 363 as genuinely faulty.

We conducted an extensive evaluation on \dataname~using ten popular LLMs, including both open-source, closed-source, and math-specialized models, to assess their capacity to function as \thinker. Our evaluation focused on three key aspects: (i) Without any hints, how effectively can LLMs recognize that a given math problem is faulty? Unfortunately, even the best-performing model, Gemini 1.5 Pro, identified only 33.33\% of the faulty problems successfully. (ii) When provided with hints—either factually correct or misleading ones, how reliably do LLMs behave as \thinker? Our results indicate that factually correct hints about the problem's faultiness significantly improve performance, but models still fall short of perfection. Moreover, LLMs demonstrated vulnerability to misleading hints, with models like Gemini and GPT-4 exhibiting confirmational bias when influenced by incorrect hints. (iii) How accurate are the explanations generated by LLMs when they correctly identify a faulty problem? Our findings reveal that models such as Gemini and Qwen-1.5 consistently produced correct explanations when they recognized falsehoods, highlighting their ability to articulate reasoning effectively in such cases.

In summary, this study advocates for the development of LLMs that go beyond mere \solver~and evolve into true \thinker. The key contributions of this paper are threefold: 
\begin{itemize}
\item We identify the critical gap in LLMs' mathematical reasoning abilities, highlighting the need to shift from \solver~to \thinker. 
\item We introduce \dataname, a balanced dataset comprising 363 faulty math problems, specifically designed to evaluate this aspect of LLM performance. 
\item Through extensive analysis, we provide comprehensive insights into the strengths and limitations of current LLMs in mathematical reasoning and their capacity to detect logical inconsistencies. 
\end{itemize}

\section{Related Work}

Mathematical problem-solving has long been a significant area of AI research. One of the pioneering works in automatic math word problem (MWP) solving \citep{bobrow1964question} developed a system that used rules and a parser to translate word problems into algebraic equations for computational solving.
With the advent of deep learning, Seq2Seq models \citep{wang2017deep,huang2018neural,saxton2018analysing,wang2019template,xie2019goal,liang2022mwp} and Seq2Tree \citep{wang2018translating,liu2019tree,wu2020knowledge} models began leveraging neural network architectures to process and solve complex mathematical problems, laying the foundation for further advancements in the field. 
Recently, we have witnessed the development of general-purpose LLMs such as GPT-4, Claude3 \citep{anthropic2024claude}, Gemini \citep{reid2024gemini}, Llama3 \citep{meta2024llama3}, etc. Mathematical reasoning benchmarks GSN8K \citep{cobbe2021training} and MATH \citep{hendrycksmath2021}, have become crucial in the evaluation of LLMs' math problem-solving capability.
In addition, there are specific LLMs trained on mathematical corpora \citep{azerbayev2021llemma,wang2023generative,paster2023openwebmath,toshniwal2024openmathinstruct}, including MathVerse \citep{lewkowycz2022solving}, Internlm-Math \citep{ying2024internlm} and DeepSeekMath \citep{shao2024deepseekmath}.

\begin{figure*}[t]
\centering
\includegraphics[width=1.0\linewidth]{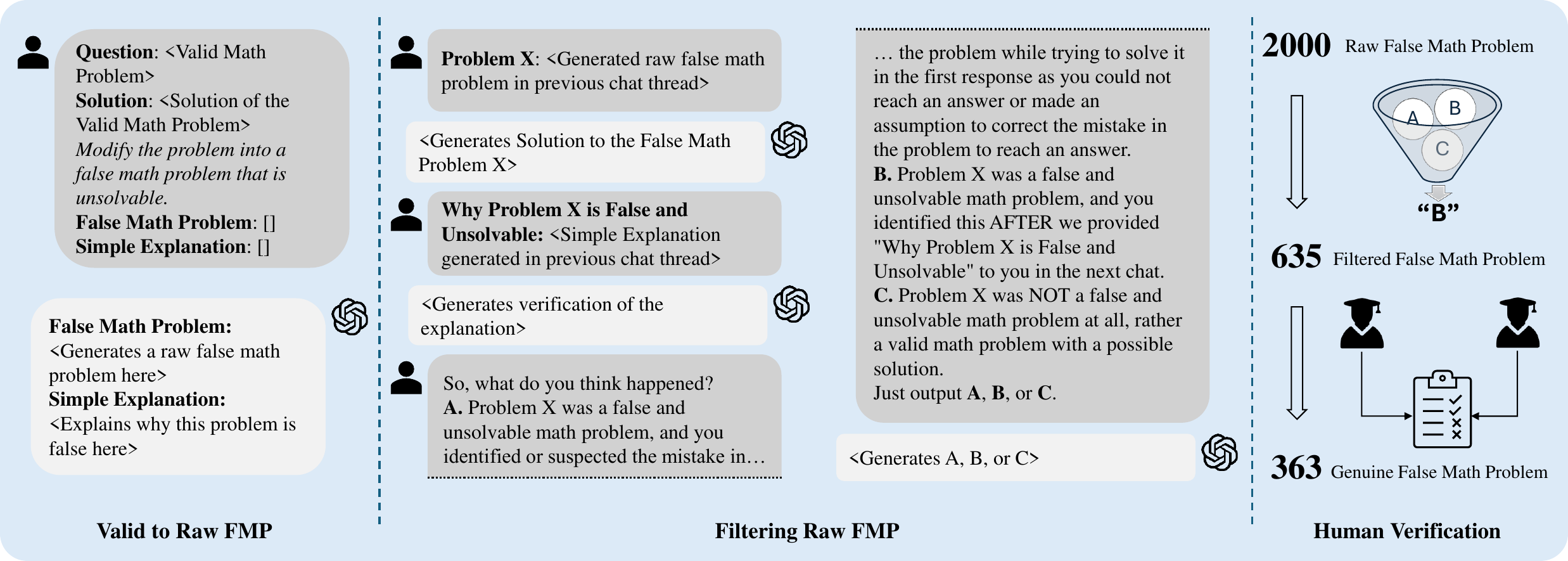} 
\caption{\dataname~curation process. Three stages: i) GPT-4 converts valid math problems into faulty ones; ii) GPT-4 self-verifies; iii) human verification.}
\label{fig:dataset_creation}
\end{figure*}

Several techniques have been developed to improve the LLM performance on those benchmarks. Prompt engineering, including zero-shot and few-shot learning \citep{brown2020language,lewkowycz2022solving}, involves designing specific prompts to guide the model to produce more accurate and contextually relevant outputs. Chain-of-thought (CoT) prompting \citep{wei2022chain} is another technique that involves converting a query into a set of intermediate queries in a sequential manner to enhance the reasoning process of LLMs. Deeply Understanding the Problems (DUP) \citep{zhong2024achieving} emphasizes a thorough comprehension of the problem before attempting to solve it, ensuring that the nuances and complexities are fully grasped. Program-aided language modeling \citep{zhou2023solving,zhang2024evaluating,ying2024internlm} leverages the code completion capabilities of language models to enhance their reasoning capacities, as demonstrated in recent studies. Agent-aided approaches \citep{lei2024macm,toshniwal2024openmath}, including multi-agent techniques, collectively contribute to significant improvements in the mathematical reasoning and problem-solving abilities of modern LLMs.

However, prior work has primarily focused on enhancing LLM performance on valid math problems, leaving the logical reasoning abilities of LLMs in more complex mathematical environments largely underexplored. Our study distinguishes LLMs' roles as \thinker~from \solver, and introduces a collection of faulty math problems, both with and without hints, to evaluate their true capacity for mathematical reasoning.

\section{\dataname~Curation and Statistics}

Generating faulty math problems is not only inherently a creative process, but it is also unconventional, as it is not a typical human task.
Our initial approach involved enlisting American high school math competition (AMC) participants to create faulty math problems from scratch. However, this effort resulted in a problem set that is biased toward particular mathematical categories and specific types of falsehoods. We then explored data augmentation via LLMs, using this seed dataset created by high-school students, but this approach also failed. Inspired by recent studies on generating synthetic data using LLMs \citep{DBLPLou0XSAX0024,Yuan2023}, we extensively experimented with modifying valid problems from the MATH dataset using GPT-4 by giving it well-crafted prompts. This approach significantly improved the likelihood of generating genuinely false math problems.
We began by generating a pool of raw false math problems using GPT-4. After a carefully designed filtering process conducted by GPT-4, human evaluators assessed the falsehood of these problems. The entire process is summarized in Figure \ref{fig:dataset_creation}, with details provided in the following section.


\subsection{\dataname~Curation}
\paragraph{Valid to Raw Faulty Math Problem.}
To compile a diverse collection of faulty math problems across different categories (e.g., Algebra, Geometry), we used the MATH dataset \citep{hendrycks2021measuring} as our source for valid problems. We selected 400 problems from each of five categories—Algebra, Geometry, Counting \& Probability, Number Theory, and Precalculus—totaling 2,000 valid problems. Given its creativity and strong performance on the MATH dataset, we chose GPT-4 as our LLM for this stage.

We provided GPT-4 with each valid math problem and its correct solution, instructing it to 
\begin{itemize}
    \item convert the problem into a false and unsolvable version, and
    \item generate a brief explanation of why the problem is faulty—a crucial step that significantly improved the generation of genuinely faulty problems.
\end{itemize}

\paragraph{LLM Self-Filtering.}
We found that many of the 2,000 generated false math problems were actually solvable, so we implemented a self-verification process to filter out those that did not meet the criteria for being truly unsolvable.

In this self-evaluation process, GPT-4 was initially presented with a ``raw false math problem'' to solve. After providing a solution, it was prompted with a ``simple explanation" of why the problem was unsolvable, encouraging GPT-4 to reflect on its initial solution. Finally, we posed a multiple-choice question to GPT-4, asking for its final decision on the entire conversation. The choices (A, B, or C) are outlined in Figure \ref{fig:dataset_creation}.

If GPT-4 outputs ``A", it implies that GPT-4 believes the raw faulty math problem to be a genuine faulty math problem but not a challenging one to identify. If GPT-4 outputs ``B", it implies that GPT-4 considered the raw false math problem to be a genuine false math problem and a challenging one that it could not identify to be false when initially presented. An output of ``C" implies that the false math problem is a valid math problem with a possible solution.

We filtered out the problems marked ``C" and ``A," as they had a lower likelihood of being genuine and challenging false math problems. This left us with 635 problems.

\paragraph{Human Verification.}
The remaining 635 math problems were then reviewed by human annotators for verification. Each annotator independently assessed all 635 problems to determine whether they were genuinely false, meaning no solution existed as presented. Through this manual verification process, 363 problems were confirmed as genuine false math problems by both annotators. The other 272 problems were found to be solvable and were discarded. 

\subsection{\dataname~Statistics.} 

\begin{figure}[t]
    \centering
    \includegraphics[width=1.0\textwidth]{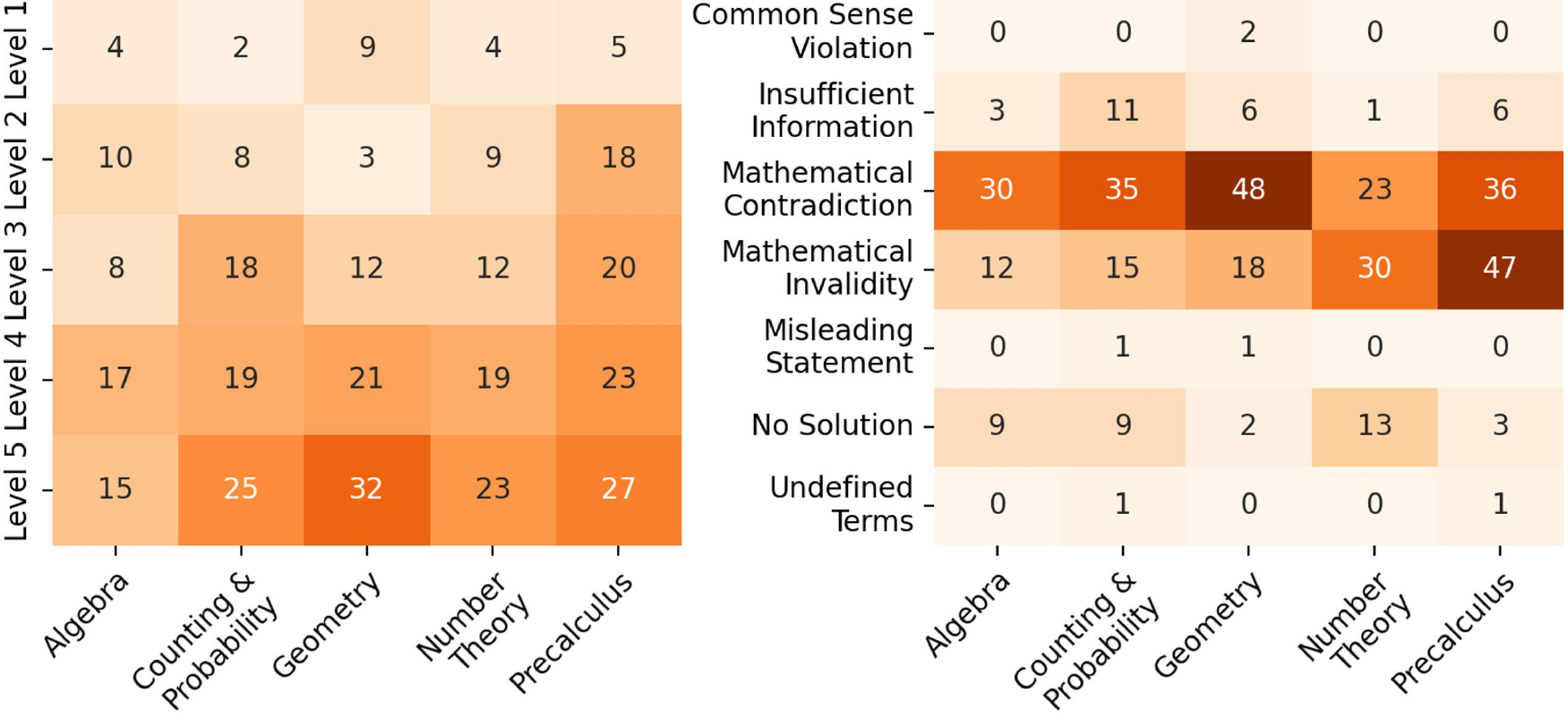}
    \caption{Distributions of the dataset by difficulty level and math category (left), and by the origin of falsehood (right).}
    \label{fig:dataset}
\end{figure}

Figure \ref{fig:dataset} (left) illustrates the distribution of false math problems across various mathematical categories, demonstrating GPT-4's balanced capability in generating false math problems for each category. A notable trend within the dataset is the prevalence of higher-difficulty problems, indicating that more complex math problems are more likely to mislead GPT-4 when transformed into false problems.

To understand the origins of errors in \dataname, we randomly selected 50 samples and manually classified the types of falsehoods. Subsequently, we used GPT-4o to systematically categorize each false problem into these predetermined categories. As shown in Figure \ref{fig:dataset} (right), the analysis reveals that \emph{Mathematical Contradiction} and \emph{Invalidity} are the most common types of errors generated by GPT-4. This trend suggests that the model tends to introduce these specific errors during the problem conversion process because these two types are easier to construct and are applicable across various problems. Additionally, these types of errors are more likely to be retained as they effectively challenge the model's problem-solving accuracy. In contrast, errors such as \emph{Common Sense Violation}, \emph{Misleading Statements}, and \emph{Undefined Terms} occur less frequently, showcasing the model's preferences in generating false math problems. The analysis highlights the model's ability to generate complex false problems and its tendency towards certain types of errors, showcasing both the strengths and preferences of the model.

\section{Evaluation Method}
The primary task for LLMs in this research is to determine whether they can identify issues in given false math problems. However, directly asking this question would signal to the LLMs that the problem is false, thereby influencing their response. The same constraint applies to human annotators; we cannot explicitly ask whether the questions are false. Instead, we ask them to solve the problems without providing any hints. By analyzing their solutions, we can determine whether the falsehood was detected. For instance, if a solution mentions that the problem is solvable only with additional constraints, it indicates awareness of the issue. Consequently, evaluating the LLMs' responses to false math problems requires careful and nuanced assessment.

To tackle this challenge, we sought to \textbf{develop an accurate automated evaluator using an LLM, designed to closely align with human assessments}. We first established a gold standard by having two PhD-level reviewers evaluate responses from GPT-4-Turbo (2024-04-09 Update) on the dataset. The reviewers found that GPT-4-Turbo failed to identify issues in 300 of the 363 genuine false math problems, achieving a gold-standard accuracy of 17.36\% (63 out of 363 identified). We then tested different auto-evaluation prompts with two other LLMs, GPT-4 and GPT-4o, comparing their performance against the gold standard. We use $LLM_s$ to denote the LLM to be evaluated (i.e., the problem solver) and $LLM_e$ to denote the LLM as an evaluator.

\begin{table}[htbp]

\centering
\small
\begin{tabular}{lcccc}
\hline\\ [-1.5ex]
\vspace{0.1cm}\textbf{Prompt} & \textbf{LLM\(_{e}\)} & \makecell{\textbf{Acc}} & \makecell{\textbf{F1}} & \makecell{\textbf{Eval}}\\
\hline\\[-1.5ex]
\vspace{0.1cm}\makecell[l]{Human-Evaluation\\(Gold Standard)} & - & 100 & 100 & 17.36\\
\\[-1.5ex]
\vspace{0.1cm}\makecell[l]{Simple Prompt} & GPT-4o & 85.67 & 61.79 & 4.68 \\
\\[-1.5ex]
\vspace{0.1cm}\makecell[l]{Simple Prompt\\ + Explanation} & GPT-4o & 84.02 & 59.26 & 3.58 \\
\\[-1.5ex]
\vspace{0.1cm}\makecell[l]{Chain-of-thought} & GPT-4o & 85.95 & 64.57 & 4.96 \\
\\[-1.5ex]
\vspace{0.1cm}\makecell[l]{Simple Prompt} & GPT-4 & 88.15 & \textbf{80.44} & \textbf{19.83} \\
\\[-1.5ex]
\vspace{0.1cm}\makecell[l]{Simple Prompt\\ + Explanation} & GPT-4 & \textbf{89.81} & 77.99 & 9.37\\
\hline
\end{tabular}
\caption{Automatic-Evaluation Methods' performance considering Human-Evaluation as the ground truth. Inferred model: GPT-4-Turbo. Eval = the performance of GPT-4-Turbo using respective evaluation method.}
\label{table:auto_evaluation_performance}
\end{table}

\paragraph{Auto-Evaluation Prompts.}
For the automatic evaluation with $LLM_e$, we tried the following prompting techniques:
\begin{itemize}
    \item \emph{Simple Prompt}: Problem X: {[`False Math Problem’]}
LLM Y’s Solution: {[`$LLM_i$ Solution’]}
Problem X is a false math problem and cannot be solved without modifications to the problem statement. Did LLM Y identify this in its solution?
Please answer `Yes' or `No'.
    \item \emph{Simple Prompt + Explanation}: Insert ``Why Problem X is a False and Unsolvable Math: {[`Simple Explanation']}" right after the LLM Y's Solution.
    \item \emph{Chain-of-thought prompt}: Provide one positive and one negative example, and explain step by step how the LLM Y identified or missed the falsehood of the math problem. Here, instead of answering ``yes" or ``no" directly, we prompted it to reason, and finally answer.

\end{itemize}

\begin{table*}[ht]

\centering
\begin{tabular}{lcc|c|c|cc}

\hline\\ [-1.5ex]
\multirow{ 2}{*}{\textbf{Model}} & \multirow{ 2}{*}{\textbf{Scope}} & \multirow{ 2}{*}{\textbf{Source}} & \textbf{\textsc{Math}} & \textbf{w/o } & \multicolumn{2}{c}{\textbf{w/ hints}}  \\
& & & (valid math) &  \textbf{hints} & \textbf{correct} & \textbf{misleading} \\
\hline\\ [-1.5ex]
Gemini-1.5-Pro & General & Closed & 58.5 & 33.33 & 95.87 & 33.61\\
GPT-4 & General & Closed & 52.9& 31.40 & 63.36 & 28.10\\
Qwen-1.5-72B & General & Open & 41.4& 29.48 & 63.36 & 28.65 \\
Claude-3-Opus & General &  Closed &60.1&  27.82 & 86.78 & 26.45\\
Mixtral-8X22B-V0.1 & General & Open & 41.8 & 23.69 & 75.21 & 30.30 \\
Deepseek-V2 & General & Open & 43.6& 17.36 & 64.19 & 22.59\\
Yi-1.5-34B & General & Open & 50.1&  11.85 & 70.52 & 24.52\\
Deepseek-Math-7B-RL & Math & Open & 51.7&  10.19 & 69.42 & 9.37\\
Llama-3-70B & General &  Open &  50.4& 7.16 & 79.06 & 13.77\\
Internlm2-Math-20B & Math & Open &37.7& 4.96 & 58.95 & 10.19\\
\hline
\end{tabular}
\caption{Evaluation of various LLMs on the \dataname~(size: 363) when hints are unavailable (w/o hint) or not (w/ hint). We also provide their performance on \textsc{Math} \citep{hendrycks2021measuring} as reference. Auto-evaluation model: GPT-4. Metric: accuracy}
\label{table:model_evaluation}
\end{table*}

As a binary classification task with human evaluation as the gold standard (Positive: Correctly identified the false math problem; Negative: Did not identify the false math problem), the confusion matrix elements (TP, TN, FP, FN) are defined by the agreement between the auto-evaluation and the human evaluation. 

Here, we look at the F1 Measure, Accuracy, and the performance of $LLM_s$ using $LLM_e$ compared to human evaluation. The evaluation by $LLM_e$ is defined as the percentage of false math problems identified by $LLM_s$ (True Positives + False Positives). From Table \ref{table:auto_evaluation_performance}, we observe that the ``Simple Prompt + Explanation" with GPT-4 as $LLM_e$ achieves the highest correlation with human evaluation. However, the ``Simple Prompt" with GPT-4 performs best across all metrics and aligns most closely with human evaluation. Therefore, we selected the \textbf{Simple Prompt with GPT-4} as our automatic evaluator for the false math problem dataset across all LLMs in subsequent experiments.

\section{Experiments}


\subsection{Experimental Setup}

\paragraph{Large Language Models (i.e., $LLM_s$).} In this work, we evaluate three categories of state-of-the-art LLMs, known for their mathematical problem-solving abilities, with the goal of ensuring a comprehensive and representative assessment.
\begin{itemize}
    \item Closed-source LLMs: GPT-4 \citep{achiam2023gpt}, Claude 3 Opus \citep{anthropic2024claude}, and Gemini 1.5 Pro \citep{reid2024gemini}
    \item Open-source LLMs: Llama 3 70B \citep{meta_llama_3_2024}, Yi 1.5 34B \citep{young2024yi}, Mixtral 8x22B \citep{mistral2024mixtral}, Qwen 1.5 72B \citep{qwen}, and DeepSeek v2 \citep{deepseekv2}
    \item Math-specific LLMs: DeepSeek Math RL \citep{shao2024deepseekmath} and InternLM2-Math \citep{ying2024internlmmath}
\end{itemize}

In our experiments, we employed the greedy decoding strategy for all LLMs. We accessed the capabilities of the closed-source models via APIs. For the open-source models, we utilized the Hugging Face library on A100 GPUs to ensure optimal configuration. For other settings, including system prompts, we adhered strictly to the configurations specified in the respective models' papers or technical reports.

\subsection{Results and Analyses}
We try to answer the following five questions to better understand the behavior of various LLMs on \dataname.

\begin{figure}[t]
    \centering
    \includegraphics[width=1.0\textwidth]{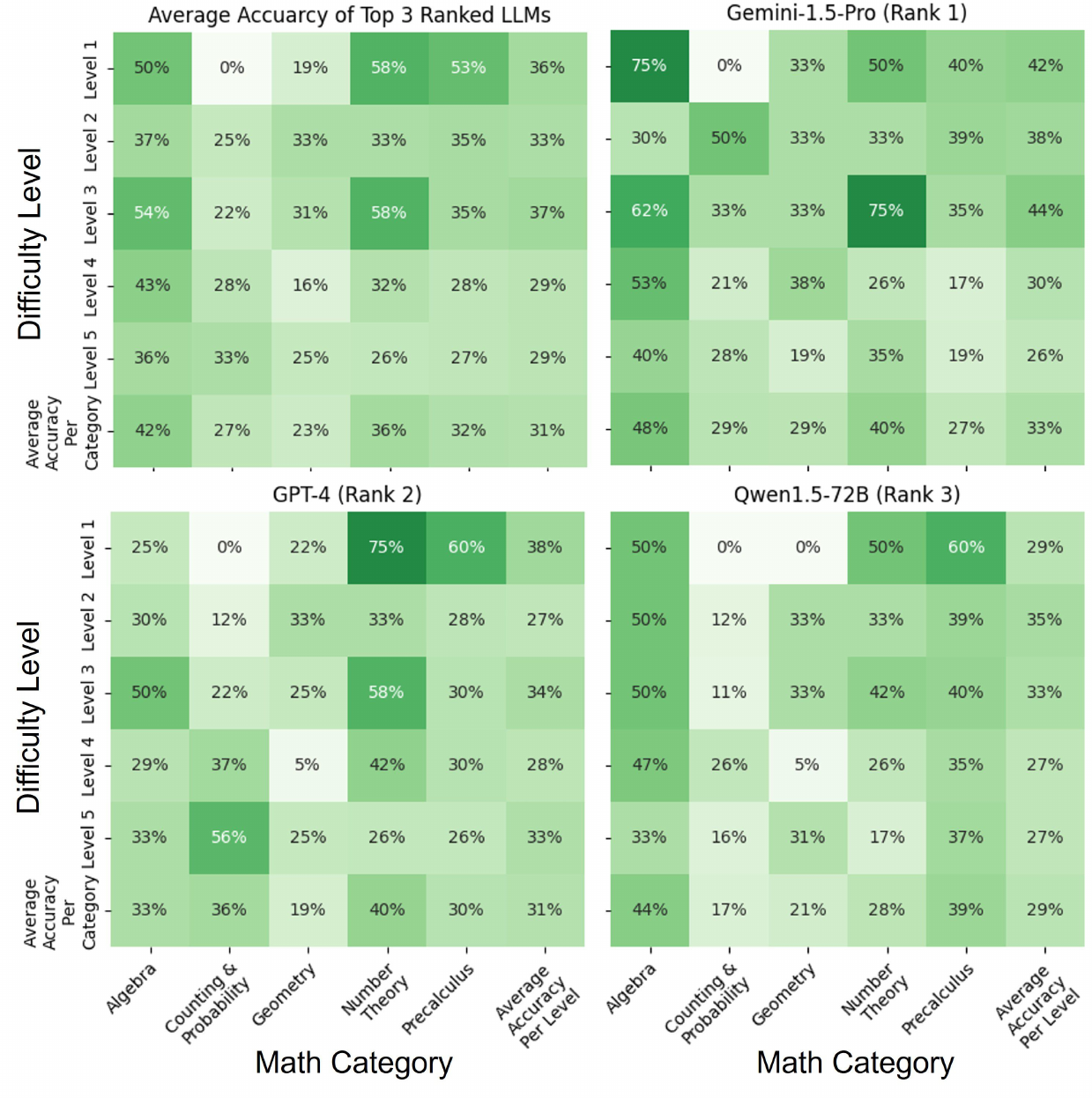}
    \caption{Accuracy heatmaps for different difficulty levels and categories. Top-left: Average accuracy of top 3 LLMs. Top-right: Gemini-1.5-Pro (Rank 1). Bottom-left: GPT-4 (Rank 2). Bottom-right: Qwen1.5-72B (Rank 3).}
    \label{fig:acc_per_level_per_category}
\end{figure}


\begin{figure}[t]
    \centering
    \includegraphics[width=0.95\textwidth]{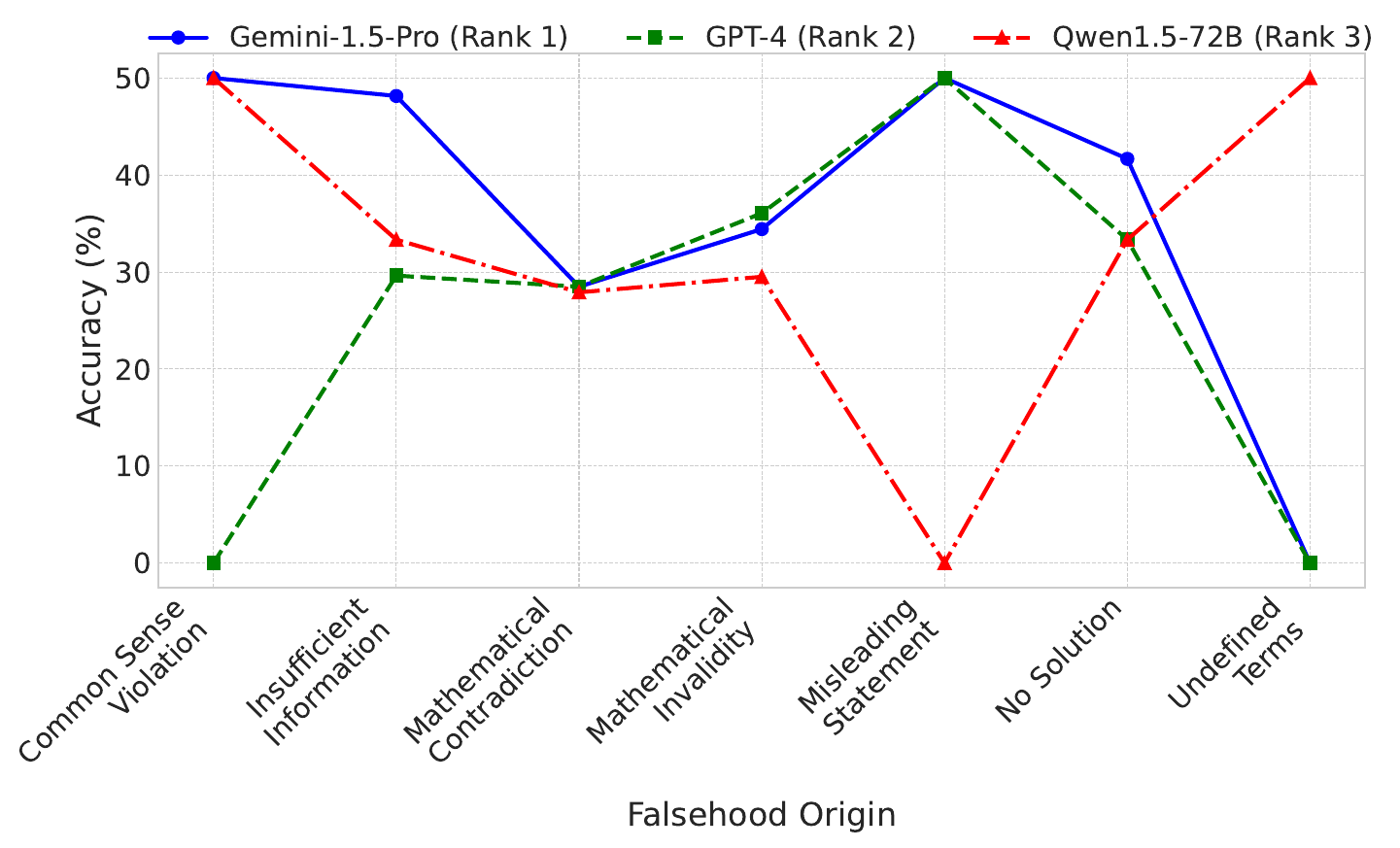}
    \caption{Accuracy of the top three LLMs across different origins of falsehood.}
    \label{fig:accuracy_by_falsehood}
\end{figure}

\paragraph{$\mathcal{Q}_1$: How do the LLMs perform  on \dataname~without any hints or with hints?} We report them separately as follows.

\textbf{When hints are not provided}, Table \ref{table:model_evaluation} (w/o hints) highlights that the Gemini-1.5-Pro model outperformed others, correctly identifying 33.33\% false problems in \dataname. Following closely were GPT-4, Qwen-1.5, and Claude-3, each demonstrating strong performance. However, models such as Llama-3, Yi-1.5, and the two math-specialized LLMs, Deepseek-Math-7B-RL and Internlm2-Math-20B, significantly underperformed in this challenge. 

By comparing performance on \dataname~with that on the \textsc{Math} dataset \citep{hendrycks2021measuring}, which consists of valid math questions, we find that strong results on general math datasets do not necessarily correlate with high effectiveness in identifying faulty math problems. This suggests that solving the faulty problems in \dataname~requires deeper reasoning and critical thinking than solving the general mathematical  problems.

\textbf{When provided hints}, we conduct this experiment where, in one round, we provide a factually correct hint to the LLM that the problem can be a false and unsolvable math problem, and in another round, we provide a misleading hint that the problem is a valid and solvable math problem whereas they are not. From Table \ref{table:model_evaluation} (w/ hints), we can infer that while LLMs score high when it has true hints, they get easily misled by the false hints. The LLMs in this experiment exhibited high confirmational bias, where they agreed to the instruction easily, without rationalizing and reaching a different conclusion.

\paragraph{$\mathcal{Q}_2$: What types and origins of faulty math questions pose the greatest challenges for LLMs?} We begin by analyzing the factor related to the type of math problem, followed by an analysis of the factor concerning the origin of the faultiness.

\enspace\enspace\enspace\textbullet\enspace\textbf{Analysis by difficulty level and mathematical category}

We conducted an in-depth analysis of the fine-grained accuracy of the top three performing LLMs (Gemini, GPT-4, Qwen1.5) across various difficulty levels (ranging from level 1 to level 5) and mathematical categories (such as Algebra, Counting \& Probability, etc.), as illustrated in Figure \ref{fig:acc_per_level_per_category}.

From the average accuracy of the top three LLMs (top-left), it is evident that the performance in identifying false math problems is slightly higher for simpler to moderately difficult problems (levels 1-3) compared to more challenging ones (levels 4-5). Particularly, compared to GPT-4 and Qwen1.5, Gemini exhibits a notable strength in this aspect. When evaluating by mathematical category, Algebra demonstrates the highest accuracy, followed by Number Theory, with Geometry being the most challenging, evidenced by an average accuracy of only 23\%.

A comparative analysis of the performance of the top three LLMs in different mathematical categories reveals varied strengths among the models. Gemini excels in Algebra and Number Theory, GPT-4 performs best in Counting \& Probability and Number Theory, while Qwen1.5 shows superior performance in Algebra and Precalculus. Conversely, GPT-4 exhibits significant deficiencies in Geometry, and Qwen1.5 struggles with Counting \& Probability and Geometry. Notably, Gemini does not exhibit substantial weaknesses in any particular category.

\enspace\enspace\enspace\textbullet\enspace\textbf{Analysis by origin of falsehood}


We investigated the fine-grained accuracy of the top three LLMs across different origins of falsehood, as depicted in Figure \ref{fig:accuracy_by_falsehood}.

A horizontal comparison among the top three LLMs reveals distinct performances: GPT-4 achieves 0\% accuracy in common sense violations, Gemini and GPT-4 both struggle with Undefined Terms, achieving 0\% accuracy, whereas Qwen1.5 attains a 50\% accuracy rate in this category. Qwen1.5, however, shows a 0\% accuracy rate in Misleading Statements, while Gemini significantly outperforms the other two in the Insufficient Information category. In the most populated classes, Mathematical Contradiction and Mathematical Invalidity, all three LLMs exhibit comparable performance.

This analysis underscores the differential capabilities and limitations of each LLM across various mathematical domains, revealing that evealing that none of the fine-grained accuracies exceed 50\%. These findings indicate that there is significant room for improvement in the task of identifying false math problems, highlighting the need for further enhancement of LLM performance in this area.

\begin{table}[h!]

\centering
\small
\begin{tabular}{lccc}
\hline\\ [-1.5ex]
\vspace{0.1cm}\textbf{Model}  & \textbf{Acc(\%)} & \textbf{Acc-S(\%)}  & \textbf{Avg(\%)}\\
\hline\\ [-1.5ex]
Gemini-1.5-Pro &  56.47 & 55.92 & 56.20 \\
GPT-4 & 25.07 & 39.94 & 32.51 \\
Claude-3-Opus &  29.20 & 22.31 & 25.76 \\
Llama-3-70B &  20.66 & 27.27 & 23.97 \\
Deepseek-V2 & 16.25 & 30.30 & 23.28 \\
Internlm2-Math-20B & 31.96 & 9.37 & 20.66 \\
Yi-1.5-34B & 22.87 & 11.02 & 16.94 \\
Mixtral-8X22B-V0.1 &  13.77 & 19.01 & 16.39 \\
Qwen-1.5-72B & 16.25 & 12.40 & 14.33 \\
\vspace{0.1cm}Deepseek-Math-7B-RL & 17.08 & 7.99 & 12.53 \\
\hline
\end{tabular}
\caption{The table presents the performance of various LLMs on Evaluation-Aware Task-Instructions that expect only ``Yes" or ``No" as outputs. The Acc-S column refers to prompts where ``Yes" and ``No" have been switched.}
\label{table:yes_no_task_aware}
\end{table}

\paragraph{$\mathcal{Q}_3$: How does LLM performance compare to human performance on \dataname~without any hints?} To answer this question, we recruited 15 graduate-level students to participate in our study.
For this human benchmarking stage, we needed to ensure that the human participants did not get any hints that the problems were false math problems to keep their responses unbiased. The participants were provided with the false math questions and asked to write down their responses. Each participant was given 10 problems from our dataset to solve within 72 hours. They were strictly prohibited from using any AI tools. The participants were asked to solve 10 problems, not all 363 problems, for two main reasons. Firstly, to reduce their effort and prevent fatigue-induced errors. Secondly, to prevent students from identifying that there is a prominent pattern (all mathematical questions are false and unsolvable) in the questions they were facing. The students' responses were manually evaluated by the same evaluators that evaluated GPT-4's responses in our auto evaluation-method step to keep the evaluation consistent. 

We observed an average accuracy of 21.33\% in identifying false math problems by the human participants. This performance is 1.6\% higher than the average of our 10 LLMs, which is 19.72\% (table \ref{table:model_evaluation}). Even though the performance gap is small, it suggests that this task is comparatively easier for human beings than LLMs. However, we also need to address that this is not an apple-to-apple comparison. LLMs were solving one problem at a time whereas the human participants had 10 questions to sense a pattern of falsehood. Due to input-window limitation, we cannot expect LLMs to reason through even two questions in many cases, let alone 10. On the other hand, a single LLM, arguably, has in its possession more mathematical knowledge than a graduate-level student; what we have observed is the failure to reason from them better than humans.

\paragraph{$\mathcal{Q}_4$: How do the LLMs react when they are asked to simply output ``Yes" or ``No"?}
Traditional NLP tasks evaluate systems based on explicit task instructions, where the system knows the evaluation criteria. For example, in machine-translation tasks, if the instruction is to translate ``Gracias" from Spanish to German, a correct response would be ``Danke." In contrast, if the instruction changes to: ``Translate the following \textit{English} to German: `Gracias'", a correct response would clarify that ``Gracias" is Spanish, not English, and then provide the German equivalent, ``Danke." Essentially, evaluating the system based on the acknowledgement of the underlying issue in the task instruction is something novel we have done in our primary experiment, $\mathcal{Q}_1$. We can consider $\mathcal{Q}_1$ as an ``Evaluation-unaware" experiment where the system is not aware of the falsehood in the task instruction and not aware that it will be evaluated based on the identification of such falsehood. For $\mathcal{Q}_4$, we want to test what happens when the system \textit{is} aware of the possibility of the falsehood.

The prompt for this ``Evaluation-aware" experiment was ``\emph{Is the above math problem a false and unsolvable math problem? Output `Yes' if it is, `No' otherwise.}". Since LLMs tend to have a bias towards always outputting ``Yes" or ``No," we flipped the statement to force the LLMs to choose logically in another round of experiments with the following prompt. ``\emph{Is the above math problem a valid and solvable math problem? Output `Yes' if it is, `No' otherwise.}" Notice how, for the same sample in the dataset, we expect the LLM to respond in ``Yes" in one and ``No" in another. 

The results are detailed in Table \ref{table:yes_no_task_aware}. We can observe that, the contrast between switching ``Yes" and ``No" is very high in some LLMs. This adds to the notion that some of the LLMs have a high bias towards selecting ``Yes" or ``No" and their mathematical reasoning capacity without step-by-step generative approach is very weak.

\paragraph{$\mathcal{Q}_5$: How accurate are the explanations provided by LLMs when identifying false math problems?}

To deepen our understanding beyond binary performance evaluation, we investigate the correctness of LLMs' explanations when 
they identify a math problem as false. Specifically, we selected the top three LLMs and examined 21 samples, which all three LLMs correctly identified as false. Our analysis involved two rounds of manual verification. The first round ensured that all selected samples were indeed correctly identified as false, and the second round focused on assessing the correctness of the explanations provided. Our findings revealed that both Gemini and Qwen1.5 provided accurate explanations for all 21 samples (21/21), while GPT-4 offered correct explanations for 20 out of 21 samples (20/21). These results suggest that when an LLM identifies a math problem as false, its explanation is generally reliable.

\section{Conclusion}
This study explores the ability of LLMs in identifying false and unsolvable mathematical problems, revealing notable limitations in their logical reasoning. A major contribution of this paper is the creation of a dedicated dataset of false math problems, named \dataname, which serves as a crucial resource for evaluating and improving LLMs. The findings underscore the necessity for refining LLM training methodologies and prompt design to enhance their reliability and versatility in mathematical problem-solving, paving the way for more robust AI systems.


\clearpage
\bibliography{iclr2025_conference}

\begin{thebibliography}{41}
\providecommand{\natexlab}[1]{#1}
\providecommand{\url}[1]{\texttt{#1}}
\expandafter\ifx\csname urlstyle\endcsname\relax
  \providecommand{\doi}[1]{doi: #1}\else
  \providecommand{\doi}{doi: \begingroup \urlstyle{rm}\Url}\fi

\bibitem[Achiam et~al.(2023)Achiam, Adler, Agarwal, Ahmad, Akkaya, Aleman, Almeida, Altenschmidt, Altman, Anadkat, et~al.]{achiam2023gpt}
Josh Achiam, Steven Adler, Sandhini Agarwal, Lama Ahmad, Ilge Akkaya, Florencia~Leoni Aleman, Diogo Almeida, Janko Altenschmidt, Sam Altman, Shyamal Anadkat, et~al.
\newblock Gpt-4 technical report.
\newblock \emph{arXiv preprint arXiv:2303.08774}, 2023.

\bibitem[Ahn et~al.(2024)Ahn, Verma, Lou, Liu, Zhang, and Yin]{DBLPAhnVLLZY24}
Janice Ahn, Rishu Verma, Renze Lou, Di~Liu, Rui Zhang, and Wenpeng Yin.
\newblock Large language models for mathematical reasoning: Progresses and challenges.
\newblock In Neele Falk, Sara Papi, and Mike Zhang (eds.), \emph{Proceedings of the 18th Conference of the European Chapter of the Association for Computational Linguistics, {EACL} 2024: Student Research Workshop, St. Julian's, Malta, March 21-22, 2024}, pp.\  225--237. Association for Computational Linguistics, 2024.
\newblock URL \url{https://aclanthology.org/2024.eacl-srw.17}.

\bibitem[Anthropic(2024)]{anthropic2024claude}
AI~Anthropic.
\newblock The claude 3 model family: Opus, sonnet, haiku.
\newblock \emph{Claude-3 Model Card}, 2024.

\bibitem[Azerbayev et~al.(2021)Azerbayev, Schoelkopf, Paster, Santos, McAleer, Jiang, Deng, Biderman, and Welleck]{azerbayev2021llemma}
Z~Azerbayev, H~Schoelkopf, K~Paster, MD~Santos, S~McAleer, AQ~Jiang, J~Deng, S~Biderman, and S~Welleck.
\newblock Llemma: An open language model for mathematics.(2023).
\newblock \emph{arXiv preprint arXiv:2310.10631}, 2021.

\bibitem[Bai et~al.(2023)Bai, Bai, Chu, Cui, Dang, Deng, Fan, Ge, Han, Huang, Hui, Ji, Li, Lin, Lin, Liu, Liu, Lu, Lu, Ma, Men, Ren, Ren, Tan, Tan, Tu, Wang, Wang, Wang, Wu, Xu, Xu, Yang, Yang, Yang, Yang, Yao, Yu, Yuan, Yuan, Zhang, Zhang, Zhang, Zhang, Zhou, Zhou, Zhou, and Zhu]{qwen}
Jinze Bai, Shuai Bai, Yunfei Chu, Zeyu Cui, Kai Dang, Xiaodong Deng, Yang Fan, Wenbin Ge, Yu~Han, Fei Huang, Binyuan Hui, Luo Ji, Mei Li, Junyang Lin, Runji Lin, Dayiheng Liu, Gao Liu, Chengqiang Lu, Keming Lu, Jianxin Ma, Rui Men, Xingzhang Ren, Xuancheng Ren, Chuanqi Tan, Sinan Tan, Jianhong Tu, Peng Wang, Shijie Wang, Wei Wang, Shengguang Wu, Benfeng Xu, Jin Xu, An~Yang, Hao Yang, Jian Yang, Shusheng Yang, Yang Yao, Bowen Yu, Hongyi Yuan, Zheng Yuan, Jianwei Zhang, Xingxuan Zhang, Yichang Zhang, Zhenru Zhang, Chang Zhou, Jingren Zhou, Xiaohuan Zhou, and Tianhang Zhu.
\newblock Qwen technical report.
\newblock \emph{arXiv preprint arXiv:2309.16609}, 2023.

\bibitem[Bobrow(1964)]{bobrow1964question}
Daniel~G Bobrow.
\newblock A question-answering system for high school algebra word problems.
\newblock In \emph{Proceedings of the October 27-29, 1964, fall joint computer conference, part I}, pp.\  591--614, 1964.

\bibitem[Brown et~al.(2020)Brown, Mann, Ryder, Subbiah, Kaplan, Dhariwal, Neelakantan, Shyam, Sastry, Askell, et~al.]{brown2020language}
Tom Brown, Benjamin Mann, Nick Ryder, Melanie Subbiah, Jared~D Kaplan, Prafulla Dhariwal, Arvind Neelakantan, Pranav Shyam, Girish Sastry, Amanda Askell, et~al.
\newblock Language models are few-shot learners.
\newblock \emph{Advances in neural information processing systems}, 33:\penalty0 1877--1901, 2020.

\bibitem[Cobbe et~al.(2021)Cobbe, Kosaraju, Bavarian, Chen, Jun, Kaiser, Plappert, Tworek, Hilton, Nakano, et~al.]{cobbe2021training}
Karl Cobbe, Vineet Kosaraju, Mohammad Bavarian, Mark Chen, Heewoo Jun, Lukasz Kaiser, Matthias Plappert, Jerry Tworek, Jacob Hilton, Reiichiro Nakano, et~al.
\newblock Training verifiers to solve math word problems.
\newblock \emph{arXiv preprint arXiv:2110.14168}, 2021.

\bibitem[DeepSeek-AI(2024)]{deepseekv2}
DeepSeek-AI.
\newblock Deepseek-v2: A strong, economical, and efficient mixture-of-experts language model, 2024.

\bibitem[Hendrycks et~al.(2021{\natexlab{a}})Hendrycks, Burns, Kadavath, Arora, Basart, Tang, Song, and Steinhardt]{hendrycks2021measuring}
Dan Hendrycks, Collin Burns, Saurav Kadavath, Akhil Arora, Steven Basart, Dawn Tang, Dawn Song, and Jacob Steinhardt.
\newblock Measuring mathematical problem solving with the math dataset.
\newblock \emph{arXiv preprint arXiv:2103.03874}, 2021{\natexlab{a}}.

\bibitem[Hendrycks et~al.(2021{\natexlab{b}})Hendrycks, Burns, Kadavath, Arora, Basart, Tang, Song, and Steinhardt]{hendrycksmath2021}
Dan Hendrycks, Collin Burns, Saurav Kadavath, Akul Arora, Steven Basart, Eric Tang, Dawn Song, and Jacob Steinhardt.
\newblock Measuring mathematical problem solving with the math dataset.
\newblock \emph{NeurIPS}, 2021{\natexlab{b}}.

\bibitem[Huang et~al.(2018)Huang, Liu, Lin, and Yin]{huang2018neural}
Danqing Huang, Jing Liu, Chin-Yew Lin, and Jian Yin.
\newblock Neural math word problem solver with reinforcement learning.
\newblock In \emph{Proceedings of the 27th International Conference on Computational Linguistics}, pp.\  213--223, 2018.

\bibitem[Imani et~al.(2023)Imani, Du, and Shrivastava]{DBLPImaniD023}
Shima Imani, Liang Du, and Harsh Shrivastava.
\newblock Mathprompter: Mathematical reasoning using large language models.
\newblock In Sunayana Sitaram, Beata~Beigman Klebanov, and Jason~D. Williams (eds.), \emph{Proceedings of the The 61st Annual Meeting of the Association for Computational Linguistics: Industry Track, {ACL} 2023, Toronto, Canada, July 9-14, 2023}, pp.\  37--42. Association for Computational Linguistics, 2023.
\newblock \doi{10.18653/V1/2023.ACL-INDUSTRY.4}.
\newblock URL \url{https://doi.org/10.18653/v1/2023.acl-industry.4}.

\bibitem[Lei(2024)]{lei2024macm}
Bin Lei.
\newblock Macm: Utilizing a multi-agent system for condition mining in solving complex mathematical problems, 2024.

\bibitem[Lewkowycz et~al.(2022)Lewkowycz, Andreassen, Dohan, Dyer, Michalewski, Ramasesh, Slone, Anil, Schlag, Gutman-Solo, et~al.]{lewkowycz2022solving}
Aitor Lewkowycz, Anders Andreassen, David Dohan, Ethan Dyer, Henryk Michalewski, Vinay Ramasesh, Ambrose Slone, Cem Anil, Imanol Schlag, Theo Gutman-Solo, et~al.
\newblock Solving quantitative reasoning problems with language models.
\newblock \emph{Advances in Neural Information Processing Systems}, 35:\penalty0 3843--3857, 2022.

\bibitem[Liang et~al.(2022)Liang, Zhang, Wang, Qin, Lan, Shao, and Zhang]{liang2022mwp}
Zhenwen Liang, Jipeng Zhang, Lei Wang, Wei Qin, Yunshi Lan, Jie Shao, and Xiangliang Zhang.
\newblock Mwp-bert: Numeracy-augmented pre-training for math word problem solving.
\newblock In \emph{Findings of the Association for Computational Linguistics: NAACL 2022}, pp.\  997--1009, 2022.

\bibitem[Liu et~al.(2019)Liu, Guan, Li, and Kawahara]{liu2019tree}
Qianying Liu, Wenyv Guan, Sujian Li, and Daisuke Kawahara.
\newblock Tree-structured decoding for solving math word problems.
\newblock In \emph{Proceedings of the 2019 conference on empirical methods in natural language processing and the 9th international joint conference on natural language processing (EMNLP-IJCNLP)}, pp.\  2370--2379, 2019.

\bibitem[Lou et~al.(2024)Lou, Zhang, Xie, Sun, Ahn, Xu, Su, and Yin]{DBLPLou0XSAX0024}
Renze Lou, Kai Zhang, Jian Xie, Yuxuan Sun, Janice Ahn, Hanzi Xu, Yu~Su, and Wenpeng Yin.
\newblock {MUFFIN:} curating multi-faceted instructions for improving instruction following.
\newblock In \emph{The Twelfth International Conference on Learning Representations, {ICLR} 2024, Vienna, Austria, May 7-11, 2024}. OpenReview.net, 2024.
\newblock URL \url{https://openreview.net/forum?id=1vrS1zwekw}.

\bibitem[Meta(2024)]{meta2024llama3}
Meta.
\newblock Meta llama 3, 2024.
\newblock URL \url{https://ai.meta.com/blog/meta-llama-3/}.
\newblock Accessed: 2024-06-12.

\bibitem[{Meta AI}(2024)]{meta_llama_3_2024}
{Meta AI}.
\newblock Meta llama 3.
\newblock \url{https://ai.meta.com/blog/meta-llama-3/}, 2024.
\newblock Accessed: 22-October-2024.

\bibitem[{Mistral AI}(2024)]{mistral2024mixtral}
{Mistral AI}.
\newblock Mixtral 8x22b: The new frontier in ai models, 2024.
\newblock URL \url{https://mistral.ai/news/mixtral-8x22b/}.
\newblock Accessed: 2024-08-13.

\bibitem[Paster et~al.(2023)Paster, Santos, Azerbayev, and Ba]{paster2023openwebmath}
Keiran Paster, Marco~Dos Santos, Zhangir Azerbayev, and Jimmy Ba.
\newblock Openwebmath: An open dataset of high-quality mathematical web text.
\newblock \emph{arXiv preprint arXiv:2310.06786}, 2023.

\bibitem[Reid et~al.(2024)Reid, Savinov, Teplyashin, Lepikhin, Lillicrap, Alayrac, Soricut, Lazaridou, Firat, Schrittwieser, et~al.]{reid2024gemini}
Machel Reid, Nikolay Savinov, Denis Teplyashin, Dmitry Lepikhin, Timothy Lillicrap, Jean-baptiste Alayrac, Radu Soricut, Angeliki Lazaridou, Orhan Firat, Julian Schrittwieser, et~al.
\newblock Gemini 1.5: Unlocking multimodal understanding across millions of tokens of context.
\newblock \emph{arXiv preprint arXiv:2403.05530}, 2024.

\bibitem[Saxton et~al.(2019)Saxton, Grefenstette, Hill, and Kohli]{saxton2018analysing}
David Saxton, Edward Grefenstette, Felix Hill, and Pushmeet Kohli.
\newblock Analysing mathematical reasoning abilities of neural models.
\newblock In \emph{International Conference on Learning Representations}, 2019.
\newblock URL \url{https://openreview.net/forum?id=H1gR5iR5FX}.

\bibitem[Shao et~al.(2024)Shao, Wang, Zhu, Xu, Song, Zhang, Li, Wu, and Guo]{shao2024deepseekmath}
Zhihong Shao, Peiyi Wang, Qihao Zhu, Runxin Xu, Junxiao Song, Mingchuan Zhang, YK~Li, Y~Wu, and Daya Guo.
\newblock Deepseekmath: Pushing the limits of mathematical reasoning in open language models.
\newblock \emph{arXiv preprint arXiv:2402.03300}, 2024.

\bibitem[Toshniwal et~al.(2024{\natexlab{a}})Toshniwal, Moshkov, Narenthiran, Gitman, Jia, and Gitman]{toshniwal2024openmath}
Shubham Toshniwal, Ivan Moshkov, Sean Narenthiran, Daria Gitman, Fei Jia, and Igor Gitman.
\newblock Openmathinstruct-1: A 1.8 million math instruction tuning dataset.
\newblock \emph{arXiv preprint arXiv: Arxiv-2402.10176}, 2024{\natexlab{a}}.

\bibitem[Toshniwal et~al.(2024{\natexlab{b}})Toshniwal, Moshkov, Narenthiran, Gitman, Jia, and Gitman]{toshniwal2024openmathinstruct}
Shubham Toshniwal, Ivan Moshkov, Sean Narenthiran, Daria Gitman, Fei Jia, and Igor Gitman.
\newblock Openmathinstruct-1: A 1.8 million math instruction tuning dataset.
\newblock \emph{arXiv preprint arXiv:2402.10176}, 2024{\natexlab{b}}.

\bibitem[Wang et~al.(2018)Wang, Wang, Cai, Zhang, and Liu]{wang2018translating}
Lei Wang, Yan Wang, Deng Cai, Dongxiang Zhang, and Xiaojiang Liu.
\newblock Translating a math word problem to an expression tree.
\newblock \emph{arXiv preprint arXiv:1811.05632}, 2018.

\bibitem[Wang et~al.(2019)Wang, Zhang, Zhang, Xu, Gao, Dai, and Shen]{wang2019template}
Lei Wang, Dongxiang Zhang, Jipeng Zhang, Xing Xu, Lianli Gao, Bing~Tian Dai, and Heng~Tao Shen.
\newblock Template-based math word problem solvers with recursive neural networks.
\newblock In \emph{Proceedings of the AAAI Conference on Artificial Intelligence}, volume~33, pp.\  7144--7151, 2019.

\bibitem[Wang et~al.(2017)Wang, Liu, and Shi]{wang2017deep}
Yan Wang, Xiaojiang Liu, and Shuming Shi.
\newblock Deep neural solver for math word problems.
\newblock In \emph{Proceedings of the 2017 conference on empirical methods in natural language processing}, pp.\  845--854, 2017.

\bibitem[Wang et~al.(2023)Wang, Xia, and Liu]{wang2023generative}
Zengzhi Wang, Rui Xia, and Pengfei Liu.
\newblock Generative ai for math: Part i--mathpile: A billion-token-scale pretraining corpus for math.
\newblock \emph{arXiv preprint arXiv:2312.17120}, 2023.

\bibitem[Wei et~al.(2022)Wei, Wang, Schuurmans, Bosma, Xia, Chi, Le, Zhou, et~al.]{wei2022chain}
Jason Wei, Xuezhi Wang, Dale Schuurmans, Maarten Bosma, Fei Xia, Ed~Chi, Quoc~V Le, Denny Zhou, et~al.
\newblock Chain-of-thought prompting elicits reasoning in large language models.
\newblock \emph{Advances in neural information processing systems}, 35:\penalty0 24824--24837, 2022.

\bibitem[Wu et~al.(2020)Wu, Zhang, Fu, and Huang]{wu2020knowledge}
Qinzhuo Wu, Qi~Zhang, Jinlan Fu, and Xuan-Jing Huang.
\newblock A knowledge-aware sequence-to-tree network for math word problem solving.
\newblock In \emph{Proceedings of the 2020 conference on empirical methods in natural language processing (EMNLP)}, pp.\  7137--7146, 2020.

\bibitem[Xie \& Sun(2019)Xie and Sun]{xie2019goal}
Zhipeng Xie and Shichao Sun.
\newblock A goal-driven tree-structured neural model for math word problems.
\newblock In \emph{Ijcai}, pp.\  5299--5305, 2019.

\bibitem[Ying et~al.(2024{\natexlab{a}})Ying, Zhang, Li, Zhou, Shao, Fei, Ma, Hong, Liu, Wang, Wang, Wu, Li, Zhou, Liu, Zhang, Zhang, Yan, Qiu, Wang, Chen, and Lin]{ying2024internlmmath}
Huaiyuan Ying, Shuo Zhang, Linyang Li, Zhejian Zhou, Yunfan Shao, Zhaoye Fei, Yichuan Ma, Jiawei Hong, Kuikun Liu, Ziyi Wang, Yudong Wang, Zijian Wu, Shuaibin Li, Fengzhe Zhou, Hongwei Liu, Songyang Zhang, Wenwei Zhang, Hang Yan, Xipeng Qiu, Jiayu Wang, Kai Chen, and Dahua Lin.
\newblock Internlm-math: Open math large language models toward verifiable reasoning, 2024{\natexlab{a}}.

\bibitem[Ying et~al.(2024{\natexlab{b}})Ying, Zhang, Li, Zhou, Shao, Fei, Ma, Hong, Liu, Wang, et~al.]{ying2024internlm}
Huaiyuan Ying, Shuo Zhang, Linyang Li, Zhejian Zhou, Yunfan Shao, Zhaoye Fei, Yichuan Ma, Jiawei Hong, Kuikun Liu, Ziyi Wang, et~al.
\newblock Internlm-math: Open math large language models toward verifiable reasoning.
\newblock \emph{arXiv preprint arXiv:2402.06332}, 2024{\natexlab{b}}.

\bibitem[Young et~al.(2024)Young, Chen, Li, Huang, Zhang, Zhang, Li, Zhu, Chen, Chang, et~al.]{young2024yi}
Alex Young, Bei Chen, Chao Li, Chengen Huang, Ge~Zhang, Guanwei Zhang, Heng Li, Jiangcheng Zhu, Jianqun Chen, Jing Chang, et~al.
\newblock Yi: Open foundation models by 01. ai.
\newblock \emph{arXiv preprint arXiv:2403.04652}, 2024.

\bibitem[Yuan et~al.(2023)Yuan, Chen, Wang, Fung, Peng, and Ji]{Yuan2023}
Lifan Yuan, Yangyi Chen, Xingyao Wang, Yi~R. Fung, Hao Peng, and Heng Ji.
\newblock Craft: Customizing llms by creating and retrieving from specialized toolsets.
\newblock \emph{arXiv preprint arXiv:2309.00147}, 2023.
\newblock URL \url{https://paperswithcode.com/paper/craft-customizing-llms-by-creating-and}.

\bibitem[Zhang et~al.(2024)Zhang, Zhou, Wei, Zhao, Sha, Wang, and Wen]{zhang2024evaluating}
Beichen Zhang, Kun Zhou, Xilin Wei, Xin Zhao, Jing Sha, Shijin Wang, and Ji-Rong Wen.
\newblock Evaluating and improving tool-augmented computation-intensive math reasoning.
\newblock \emph{Advances in Neural Information Processing Systems}, 36, 2024.

\bibitem[Zhong et~al.(2024)Zhong, Wang, Xu, Liu, Ding, Du, and Tao]{zhong2024achieving}
Qihuang Zhong, Kang Wang, Ziyang Xu, Juhua Liu, Liang Ding, Bo~Du, and Dacheng Tao.
\newblock Achieving >97

\bibitem[Zhou et~al.(2023)Zhou, Wang, Lu, Shi, Luo, Qin, Lu, Jia, Song, Zhan, et~al.]{zhou2023solving}
Aojun Zhou, Ke~Wang, Zimu Lu, Weikang Shi, Sichun Luo, Zipeng Qin, Shaoqing Lu, Anya Jia, Linqi Song, Mingjie Zhan, et~al.
\newblock Solving challenging math word problems using gpt-4 code interpreter with code-based self-verification.
\newblock In \emph{The Twelfth International Conference on Learning Representations}, 2023.

\end{thebibliography}
\bibliographystyle{iclr2025_conference}

\clearpage
\appendix
\section{Case Study}
This case study explores the responses of some large language models to a false math problem. The problem involves finding two consecutive even numbers whose squares differ by 58, which is an impossible scenario due to the mathematical properties of even numbers and their squares. The study evaluates how different models, including GPT-4, Gemini-1.5-Pro, and Claude-3-Opus, respond to the problem under different conditions: without hints, with hints suggesting the problem might be false, and with misleading hints suggesting the problem is valid. The case study reveals that while language models have advanced math problem-solving capabilities, they may struggle with recognizing inherently flawed or unsolvable problems.

\subsection*{False Math Problem Description}
\subsubsection*{False Math Problem}
Two consecutive positive even numbers are each squared. The difference of the squares is 58. What is the sum of the original two numbers?
\subsubsection*{Why It is False:}
This problem is unsolvable because the difference of squares of any two consecutive even numbers, expressed as \((x+2)^2 - x^2\), simplifies to \(4x + 4\), which is always a multiple of 4. Since 58 is not a multiple of 4, there are no consecutive even numbers whose squares have a difference of 58.
\subsection*{No Hint Prompt (Q0)}
\subsubsection*{Prompt Description:}
The false math problem is directly queried to the LLM without any additional hints added.
\subsubsection*{Prompt:}
Two consecutive positive even numbers are each squared. The difference of the squares is 58. What is the sum of the original two numbers?
\subsubsection*{GPT-4 Turbo: Did Not Identify}
Let the two consecutive positive even numbers be \( n \) and \( n+2 \), where \( n \) is an even number.

The squares of these numbers are \( n^2 \) and \( (n+2)^2 \) respectively. According to the problem, the difference of these squares is 58. We can set up the equation based on the difference of squares formula:

\[
(n+2)^2 - n^2 = 58
\]

Expanding the left side, we get:

\[
n^2 + 4n + 4 - n^2 = 58
\]

Simplifying this, we have:

\[
4n + 4 = 58
\]

Subtract 4 from both sides:

\[
4n = 54
\]

Divide both sides by 4:

\[
n = 13.5
\]

However, \( n \) must be an even number, so we need to recheck our calculations. Let's re-examine the difference of squares:

\[
(n+2)^2 - n^2 = (n+n+2)(n-(n+2)) = 2n+2
\]

Setting this equal to 58:

\[
2n + 2 = 58
\]

Subtract 2 from both sides:

\[
2n = 56
\]

Divide both sides by 2:

\[
n = 28
\]

Thus, the two consecutive even numbers are 28 and 30. The sum of these numbers is:

\[
28 + 30 = 58
\]

Therefore, the sum of the original two numbers is 58.

\subsubsection*{Gemini-1.5-Pro: Did Not Identify}
\noindent

1. Set up the equation: Let the first even number be represented by \( x \).  
The next consecutive even number is \( x + 2 \).  
The problem states: \( (x + 2)^2 - x^2 = 58 \).

2. Solve the equation: Expand the equation: \( x^2 + 4x + 4 - x^2 = 58 \).  
Simplify: \( 4x + 4 = 58 \).  
Subtract 4 from both sides: \( 4x = 54 \).  
Divide both sides by 4: \( x = 13.5 \).

3. Find the two even numbers: Since we're looking for even numbers, we need to adjust slightly. Our equation led us to 13.5, which is halfway between the two even numbers.  
The two consecutive even numbers are 12 and 14.

4. Calculate the sum: \( 12 + 14 = 26 \).

\textbf{Answer:} The sum of the original two numbers is 26.
\subsubsection*{Claude-3-Opus: Did Not Identify}
Let's solve this step by step:

1) Let's call the first even number $n$. Since the numbers are consecutive even numbers, the second number will be $n+2$.

2) Now, we are told that when each number is squared, the difference of the squares is 58. We can express this as an equation:

   $(n+2)^2 - n^2 = 58$

3) Let's expand the squares:

   $(n^2 + 4n + 4) - n^2 = 58$

4) Simplify:

   $4n + 4 = 58$

5) Subtract 4 from both sides:

   $4n = 54$

6) Divide both sides by 4:

   $n = 13.5$

7) However, we started with the assumption that $n$ is an even number. The next even number after 13.5 is 14. So, the first number is 14 and the second number is 16.

8) The sum of the original two numbers is:

   $14 + 16 = 30$

Therefore, the sum of the original two numbers is 30.
\subsubsection*{GPT-4: Identified}
Let's denote the two consecutive even numbers as \( x \) and \( x + 2 \).

The problem states that the difference of their squares is 58, so we can write the equation as follows:

\[
(x+2)^2 - x^2 = 58
\]

Expanding and simplifying this equation gives:

\[
x^2 + 4x + 4 - x^2 = 58
\]
\[
\quad 4x + 4 = 58
\]
\[
\quad 4x = 54
\]
\[
\quad x = 13.5
\]

However, \( x \) must be an even number, so there seems to be a mistake in the problem. The difference of the squares of two consecutive even numbers cannot be 58.

\subsection*{``Yes/No" Prompt 1 (Q3)}
\subsubsection*{Prompt Description}
The false math problem is appended to the beginning of the prompt with a `Yes/No' question directly asking if the problem is a false math problem or not.
\subsubsection*{Prompt:}
Two consecutive positive even numbers are each squared. The difference of the squares is 58. What is the sum of the original two numbers?

\noindent
Is the above math problem a false and unsolvable math
problem? Output `Yes’ if it is, `No’ otherwise.

\subsubsection*{GPT-4: Did Not Identify}
No
\subsubsection*{Gemini-1.5-Pro: Identified}
Here's how to solve the problem:

1. Set up the equation:

Let the first even number be represented by \( x \). The next consecutive even number is \( x + 2 \). The problem translates to the equation: \( (x + 2)^2 - x^2 = 58 \).

2. Solve for \( x \):

Expand the equation: \( x^2 + 4x + 4 - x^2 = 58 \).

Simplify: \( 4x + 4 = 58 \).

Subtract 4 from both sides: \( 4x = 54 \).

Divide both sides by 4: \( x = 13.5 \).

3. Analyze the solution:

We found that \( x = 13.5 \), which is not an even number.

Conclusion:

Since the problem states we need consecutive even numbers, and our solution leads to a non-integer, the problem is unsolvable with the given conditions.

Answer: YES

\subsubsection*{Claude-3-Opus: Did Not Identify}
NO

The given problem is a valid and solvable math problem. Let's solve it step by step:

Let the two consecutive even numbers be \( x \) and \( x + 2 \).

Squaring these numbers: \( x^2 \) and \( (x + 2)^2 \).

The difference of the squares is given as 58:
\[
(x + 2)^2 - x^2 = 58
\]

Expand \( (x + 2)^2 \):
\[
x^2 + 4x + 4 - x^2 = 58
\]

Simplify:
\[
4x + 4 = 58
\]

Subtract 4 from both sides:
\[
4x = 54
\]

Divide both sides by 4:
\[
x = 13.5
\]

Since we are looking for even numbers, \( x \) must be an integer. The closest even integer to 13.5 is 14.

Therefore, the two consecutive even numbers are 14 and 16.

The sum of the original two numbers is \( 14 + 16 = 30 \).
\subsection*{``Yes/No" Prompt 2 (Q3)}
\subsubsection*{Prompt Description}
The false math problem is appended to the beginning of the prompt with a `Yes/No' question directly asking if the problem is a valid math problem or not. This is a switched question to the previous prompt, expecting the opposite response than the previous one.
\subsubsection*{Prompt:}
Two consecutive positive even numbers are each squared. The difference of the squares is 58. What is the sum of the original two numbers?

\noindent
Is the above math problem a valid and solvable math
problem? Output `Yes’ if it is, `No’ otherwise.

\subsubsection*{GPT-4: Did Not Identify}
Yes
\subsubsection*{Gemini-1.5-Pro: Identified}
Here's how to determine if the problem is valid and solvable:

1. Set up the equation:

Let the first even number be represented by \( x \). The next consecutive even number is \( x + 2 \). The problem translates to the equation: \( (x + 2)^2 - x^2 = 58 \).

2. Simplify and solve the equation:

Expand: \( x^2 + 4x + 4 - x^2 = 58 \).

Simplify: \( 4x + 4 = 58 \).

Subtract 4 from both sides: \( 4x = 54 \).

Divide both sides by 4: \( x = 13.5 \).

3. Analyze the result:

We got a solution for \( x \), but the problem states the numbers must be positive even numbers. 13.5 is not even.

Conclusion:

The problem is not solvable as it leads to a contradiction.

Output: NO

\subsubsection*{Claude-3-Opus: Did Not Identify}
YES

The given problem is a valid and solvable math problem. It provides enough information to set up an equation and solve for the two consecutive even numbers.

\subsection*{True Hint Prompt (Q4)}
\subsubsection*{Prompt Description:}
The false math problem is appended to the beginning of the prompt with a true hint indicating the likelihood of an unsolvable problem.

\subsubsection*{Prompt:}
Two consecutive positive even numbers are each squared. The difference of the squares is 58. What is the sum of the original two numbers?

\noindent
Solve the above math problem. Hint: The problem can be a false and unsolvable math problem.

\subsubsection*{GPT-4: Identified}
Let's denote the two consecutive even numbers as \( x \) and \( x+2 \).

The problem states that the difference of their squares is 58, so we can write the equation as follows:

\[
(x+2)^2 - x^2 = 58
\]

Expanding and simplifying the equation gives:

\[
x^2 + 4x + 4 - x^2 = 58
\]

\[
4x + 4 = 58
\]

\[
4x = 54
\]

\[
x = 13.5
\]

Since \( x \) must be an even number, there is no solution to this problem. Therefore, the problem is false and unsolvable.

\subsubsection*{Gemini-1.5-Pro: Identified}
Here's how to approach the problem, and why it has no solution:

1. Set up the equation:

- Let the first even number be represented by \( 2n \) (where \( n \) is any integer).
- The next consecutive even number is \( 2n + 2 \).
- The problem states: \( (2n + 2)^2 - (2n)^2 = 58 \)

2. Simplify and solve the equation:

- Expand the squares: \( 4n^2 + 8n + 4 - 4n^2 = 58 \)
- Simplify: \( 8n + 4 = 58 \)
- Subtract 4 from both sides: \( 8n = 54 \)
- Divide both sides by 8: \( n = 6.75 \)

3. The problem:

- We defined \( n \) as an integer, but our solution is a fraction. This means there are NO consecutive even integers that fit the given conditions.

Conclusion: The problem is unsolvable. There is no pair of consecutive even numbers whose squares have a difference of 58.

\subsubsection*{Claude-3-Opus: Identified}
Let's approach this step by step:

1) Let's assume the two consecutive even numbers are $n$ and $n+2$.

2) The problem states that the difference of their squares is 58. We can express this as an equation:

   $(n+2)^2 - n^2 = 58$

3) Let's expand the squares:

   $(n^2 + 4n + 4) - n^2 = 58$

4) Simplify:

   $4n + 4 = 58$

5) Subtract 4 from both sides:

   $4n = 54$

6) Divide both sides by 4:

   $n = 13.5$

7) However, $n$ is supposed to be an even number. There are no even numbers that, when squared and subtracted from the square of the next even number, give 58.

Therefore, this problem is unsolvable. There are no two consecutive even numbers that satisfy the given condition.
\subsection*{False Hint Prompt (Q4)}
\subsubsection*{Prompt Description}
The false math problem is appended to the beginning of the prompt with a false hint indicating that the problem is a valid and solvable math problem.
\subsubsection*{Prompt:}
Two consecutive positive even numbers are each squared. The difference of the squares is 58. What is the sum of the original two numbers?

\noindent
Solve the above math problem. Hint: The problem is a valid and solvable math problem.

\subsubsection*{GPT-4: Identified}
Let's denote the two consecutive even numbers as \( x \) and \( x+2 \).

According to the problem, the difference of their squares is 58. So, we can write the equation as follows:

\[
(x+2)^2 - x^2 = 58
\]

Expanding and simplifying the equation gives:

\[
x^2 + 4x + 4 - x^2 = 58
\]

\[
4x + 4 = 58
\]

\[
4x = 54
\]

\[
x = 13.5
\]

However, \( x \) must be an even number, so there seems to be a mistake. Let's go back to the original equation and try a different approach.

The difference of squares can be factored as follows:

\[
(x+2+x)(x+2-x) = 58
\]

\[
2x \cdot 2 = 58
\]

\[
4x = 58
\]

\[
x = 14.5
\]

Again, \( x \) must be an even number, so there seems to be a mistake.

Let's try a different approach. The difference of squares of two consecutive numbers is equal to the sum of the numbers. So, the sum of the two numbers is 58. 

However, the problem states that the difference of the squares is 58, not the sum of the numbers. 

So, there seems to be a mistake in the problem. The difference of the squares of two consecutive even numbers cannot be 58.

\subsubsection*{Gemini-1.5-Pro: Did Not Identify}
Here's how to solve the problem:

1. Set up the equation:

Let \( x \) be the first even number. The next consecutive even number is \( x + 2 \). The problem translates to the equation: \( (x + 2)^2 - x^2 = 58 \)

2. Solve the equation:

Expand the left side: \( x^2 + 4x + 4 - x^2 = 58 \)

Simplify: \( 4x + 4 = 58 \)

Subtract 4 from both sides: \( 4x = 54 \)

Divide both sides by 4: \( x = 13.5 \)

3. Find the two even numbers:

Since we're looking for even numbers, we made an error in our assumption. Let's rethink our approach.

Corrected Approach:

The difference between consecutive even numbers is always 2. The difference between their squares will increase as the numbers get larger. Let's test a few:

\( 2^2 - 0^2 = 4 \)

\( 4^2 - 2^2 = 12 \)

\( 6^2 - 4^2 = 20 \)

\( 8^2 - 6^2 = 28 \)

\( 10^2 - 8^2 = 36 \)

\( 12^2 - 10^2 = 44 \)

\( 14^2 - 12^2 = 52 \)

**\( 16^2 - 14^2 = 58 \)**  <-- This is our solution!

4. Calculate the sum:

The two original numbers are 14 and 16. Their sum is \( 14 + 16 = 30 \).

Answer: The sum of the original two numbers is 30.

\subsubsection*{Claude-3-Opus: Did Not Identify}
Great! Let's solve this problem step by step. We'll use algebra to find the two consecutive even numbers and then calculate their sum.

Given:
- Two consecutive positive even numbers are squared.
- The difference of the squares is 58.

Step 1: Let's express the two consecutive even numbers in terms of a variable. Let the first even number be \( x \). Then, the second even number will be \( x + 2 \).

Step 2: Express the squares of the two numbers and their difference.

First number squared: \( x^2 \)

Second number squared: \( (x + 2)^2 \)

Difference of the squares: \( (x + 2)^2 - x^2 = 58 \)

Step 3: Expand the squared term and simplify the equation.

\[
(x + 2)^2 - x^2 = 58
\]

\[
x^2 + 4x + 4 - x^2 = 58
\]

\[
4x + 4 = 58
\]

Step 4: Solve the equation for \( x \).

\[
4x = 54
\]

\[
x = \frac{54}{4}
\]

\[
x = 13.5
\]

Since we are looking for even numbers, \( x \) must be an even integer. The closest even integer to 13.5 is 14.

Step 5: Find the two consecutive even numbers.

First even number: \( x = 14 \)

Second even number: \( x + 2 = 14 + 2 = 16 \)

Step 6: Calculate the sum of the two even numbers.

\[
Sum = 14 + 16 = 30
\]

Therefore, the sum of the original two consecutive positive even numbers is 30.

\end{document}